# Metrics for Finite Markov Decision Processes

**Norm Ferns**
School of Computer Science
McGill University
Montréal, Canada, H3A 2A7
nferns@cs.mcgill.ca

**Prakash Panangaden**
School of Computer Science
McGill University
Montréal, Canada, H3A 2A7
prakash@cs.mcgill.ca

**Doina Precup**
School of Computer Science
McGill University
Montréal, Canada, H3A 2A7
dprecup@cs.mcgill.ca

## Abstract

We present metrics for measuring the similarity of states in a finite Markov decision process (MDP). The formulation of our metrics is based on the notion of bisimulation for MDPs, with an aim towards solving discounted infinite horizon reinforcement learning tasks. Such metrics can be used to aggregate states, as well as to better structure other value function approximators (e.g., memory-based or nearest-neighbor approximators). We provide bounds that relate our metric distances to the optimal values of states in the given MDP.

## 1 Introduction

Markov decision processes (MDPs) offer a popular mathematical tool for planning and learning in the presence of uncertainty (Boutilier et al., 1999). MDPs are a standard formalism for describing multi-stage decision making in probabilistic environments. The objective of the decision making is to maximize a cumulative measure of long-term performance, called the *return*. Dynamic programming algorithms, e.g., value iteration or policy iteration (Puterman, 1994), allow us to compute the optimal expected return for any state, as well as the way of behaving (policy) that generates this return. However, in many practical applications, the state space of an MDP is simply too large, possibly even continuous, for such standard algorithms to be applied. A typical means of overcoming such circumstances is to partition the state space in the hope of obtaining an "essentially equivalent" reduced system. One defines a new MDP over the partition blocks, and if it is small enough, it can be solved by classical methods. The hope is that optimal values and policies for the reduced MDP can be extended to optimal values and policies for the original MDP.

Recent MDP research on defining equivalence relations on MDPs (Givan et al., 2003) has built on the notion of strong probabilistic bisimulation from concurrency theory.

Bisimulation was introduced by Larsen and Skou (1991) based on ideas of Park (1981) and Milner (1980). Roughly speaking, two states of a process are deemed equivalent if all the transitions of one state can be matched by transitions of the other state, and the results are themselves bisimilar. The extension of bisimulation to transition systems with rewards was carried out in the context of MDPs by Givan, Dean and Greig (2003) and in the context of performance evaluation by Bernardo and Bravetti (2003). In both cases, the motivation is to use the equivalence relation to aggregate the states and get smaller state spaces. The basic notion of bisimulation is modified only slightly be the introduction of rewards.

The notion of equivalence for stochastic processes is problematic to use in practice because it requires that the transition probabilities agree *exactly*. This is not a robust concept, especially considering that usually, the numbers used in probabilistic models come from experimentation or are approximate estimates. A small change in probability estimates can cause bisimilar states to appear non-bisimilar. Dean, Givan and Leach (1997) addressed this issue by allowing the state space to be partitioned into blocks of states such that the states within a block are "close" in terms of their transition probabilities. However, their technique involves moving to a slightly generalized model, namely, the bounded-parameter MDP.

In this paper we address the same problem in a different way, by developing *metrics*, or distance functions, on the states of an MDP. Unlike an equivalence relation, a metric may vary smoothly as a function of the transition probabilities. Yet, a metric can be used to aggregate states in a manner similar to an equivalence relation. For example, we can choose a tolerance parameter, ε, and cluster together states that are in ε-neighborhoods. A metric can have broader applicability to other classes of function approximators as well. For instance, a metric can be used in a nearest-neighbor approximator in order to decide on the data points to be used as prototypes.

The metrics we develop are based on the notion of bisimulation. More precisely, we will require that if one of our metrics assigns a distance of 0 to a pair of states, then those states have to be bisimilar. Thus, our metrics provide a quantitative analogue of bisimulation. Additionally, our metrics will possess the following pleasing property: if the system parameters of two bisimilar states are perturbed slightly, then the two states will remain "close" in metric distance. We build on previous work by Desharnais, Panangaden, Jagadeesan and Gupta (Desharnais et al., 1999; Desharnais et al., 2002) and by van Breugel and Worrell (2001), in which the theory of bisimulation, metrics and approximation was developed for labeled Markov processes with continuous state spaces. Their work was developed in the context of formal verification; here we take the first steps to apply and extend their results in the context of optimization problems. Although we present our work currently in the context of discrete MDPs, our recent research indicates that the results can be extended for continuous MDPs.

The paper is organized as follows. Sections 2 and 3 provide the definitions and theoretical results required to construct our metrics. In section 4 we introduce two kinds of bisimulation metrics and in section 5 provide bounds on the optimal value function of MDPs that can be obtained by using these metrics for state aggregation. In section 6 we provide some experimental results to compare and contrast our metrics. Section 7 contains conclusions and directions for future work.

## 2 Background

A *finite Markov decision process* consists of a finite set of states, $S$, a finite set of actions, $A$, and for every pair of states $s$ and $s'$ and action $a$, a Markovian state transition probability, $P_{ss'}^a$, and a numerical reward, $r_s^a$. In the rest of this work we will focus on a fixed, known MDP. Moreover, since rewards are necessarily bounded, we will assume without loss of generality that $\forall a \in A, \forall s \in S.\ r_s^a \in [0,1]$.[1] We will now review briefly some basic definitions and results from MDP theory (e.g., (Puterman, 1994), sec.6.1-6.3).

A way of behaving or *policy* is defined as a mapping from states to actions, $\pi : S \to A$, and $s \in S$. The *value* of a state $s$ under policy $\pi$, $V^\pi(s)$, is defined as: $V^\pi(s) = E[\sum_{t=0}^\infty \gamma^t r_t | s_0 = s, \pi]$, where $s_0$ is the state at time $0$, $\gamma \in (0,1)$ is a discount factor for future rewards, $r_t$ is the reward obtained at time $t$, and the expectation is achieved by following the state dynamics induced by $\pi$. The mapping $V^\pi : S \to \mathbb{R}$ is called the *value function* according to $\pi$. The goal of decision making in an MDP is to find a policy $\pi$ that maximizes $V^\pi(s)$ for each $s \in S$. Such a maximizing policy and its associated value function are said to be *optimal*. Note that while there may be many optimal policies, the optimal value function, $V^*$, is unique and satisfies a family of fixed point equations,

$$V^*(s) = \max_{a \in A}(r_s^a + \gamma \sum_{s' \in S} P_{ss'}^a V^*(s')) \ \forall s \in S.$$

These are known as the *Bellman optimality equations*. They lead to the following theorem, which expresses $V^*$ as the limit of a sequence of iterates.

**Theorem 2.1.** *Let* $V_0(s) = 0$ *and*

$$V_{n+1}(s) = \max_{a \in A(s)}(r_s^a + \gamma \sum_{s' \in S} P_{ss'}^a V_n(s'))$$

*Then* $\{V_n\}_{n \in \mathbb{N}}$ *converges to* $V^*$ *uniformly.*

These results can be realized via a dynamic programming (DP) algorithm that computes a value function up to a prescribed degree of accuracy. For example, if one is given a positive tolerance $\varepsilon$ then iterating until the maximum difference between consecutive iterates is $\frac{\varepsilon(1-\gamma)}{2\gamma}$ guarantees that the current iterate differs from the true value function by at most $\varepsilon$.

Unfortunately, it is sometimes the case that the state space is too large for DP to be feasible. A standard strategy is to approximate the given MDP by aggregating its state space. The hope is that one can obtain a smaller "equivalent" MDP, with an easily computable value function, that could provide information about the value function of the original MDP. Givan, Dean, and Greig (2003) investigated several notions of state equivalence and determined that the most appropriate is stochastic bisimulation:

**Definition 2.2.** A *stochastic bisimulation relation* is an equivalence relation $R$ on $S$ that satisfies the following property:

$$sRs' \Leftrightarrow \forall a \in A.\ (r_s^a = r_{s'}^a \text{ and } \forall C \in S/R.\ P_s^a(C) = P_{s'}^a(C))$$

where $S/R$ is the state partition induced by $R$ and $P_s^a(C) = \sum_{c \in C} P_{sc}^a$. Stochastic bisimulation, $\sim$, is the largest stochastic bisimulation relation.

In (Givan et al., 2003) it was shown that the stochastic bisimulation (henceforth simply "bisimulation") partition could be found by iteratively refining partitions based on rewards and equivalence class transition probabilities, beginning with an initial partition in which all states are lumped together. This could be done in $O(|A||S|^3)$ operations.

Unfortunately, bisimulation is too stringent. Consider the sample MDP in figure 1 with 4 states labeled $s$, $t$, $u$, and $v$, and one action labeled $a$. Suppose $r_v^a = 0$. Then all states are bisimilar, because they share the same immediate reward and transition among themselves w.p.1. On the other hand, if $r_v^a > 0$ then $v$ is the only state in its bisimulation class since it is the only one with a positive reward. Moreover, $s$ and $t$ are bisimilar iff they share the same probability of transitioning to $v$'s bisimulation class. Each is bisimilar to $u$ iff that probability is zero. Thus, $u, s, t \not\sim v$, $s \sim t \iff p = q$; $s \sim u \iff p = 1.0$, and $t \sim u \iff q = 1.0$.

---

[1] If rewards are bounded between $R_{min}$ and $R_{max}$, we can achieve this by subtracting $R_{min}$ from all rewards and dividing by $R_{max} - R_{min}$

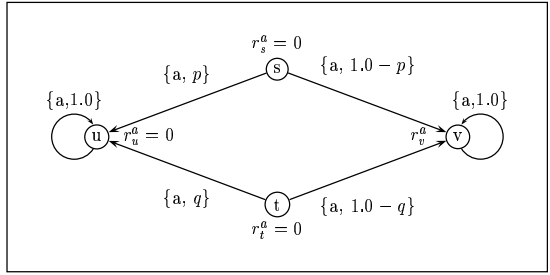

Figure 1: Sample MDP

This example demonstrates that bisimulation is simply too strong a notion; if $r_v$ is just slightly positive, and $p$ differs only slightly from $q$ then we should expect $s$ and $t$ to be practically bisimilar. From the point of view of the value function, these states will also be very close, and one can argue that aggregating them would be "safe". However, such a fine distinction cannot be made using bisimulation alone. Therefore, we seek a quantitative notion of bisimulation so that we can obtain a measure of "how bisimilar" two states are. To formulate such a notion we use semimetrics, distance functions on the state space.

**Definition 2.3.** A *semimetric* on $S$ is a map $d : S \times S \to [0, \infty)$ such that for all $s, s', s''$:

1. $s = s' \Rightarrow d(s, s') = 0$
2. $d(s, s') = d(s', s)$
3. $d(s, s'') \leq d(s, s') + d(s', s'')$

If the converse of the first axiom holds as well, we say $d$ is a *metric*.

Let $\mathcal{M}$ be the set of all semimetrics on $S$ that assign distances of at most 1. Note that every semimetric $d$ induces an equivalence relation, $R_d$, on $S$, obtained by equating points assigned distance zero by $d$.

**Definition 2.4.** We say that $d \in \mathcal{M}$ is a *bisimulation relation metric* if $R_d$ is a bisimulation relation. We say that $d$ is a *bisimulation metric* if $R_d$ is $\sim$.

## 3 Probability metrics

Our goal is to construct a class of bisimulation metrics for use in MDP state aggregation. Specifically, such metrics would be required to be easily computable and provide information concerning the optimal values of states. However, if we denote by $\mathbb{I}_{[X \not\sim Y]}$ the bisimulation metric that assigns distance 1 to states that are not bisimilar then it is not hard to show that $\mathbb{I}_{[X \not\sim Y]}$ satisfies both requirements, while possessing no more distinguishing power than that of bisimulation itself. So we additionally require that metric distances vary smoothly and proportionally with differences in rewards and differences in probabilities. Formally, we will construct bisimulation metrics via a metric on rewards and a metric on probability functions. The choice of metric on rewards is an obvious one: we simply use the absolute value of the difference. However, there are many ways of defining useful probability metrics (Gibbs & Su, 2002). Two of the most important are the Kantorovich metric and the total variation metric.[2]

Given $d \in \mathcal{M}$, the Kantorovich metric, $T_K(d)$, applied to state probability functions $P$ and $Q$ is defined by the following linear program:

$$\max_{u_i, i=1\ldots|S|} \sum_{i=1}^{|S|} (P(s_i) - Q(s_i)) u_i$$

subject to: $\forall i, j. \; u_i - u_j \leq d(s_i, s_j)$
$\forall i. \; 0 \leq u_i \leq 1$

which is equivalent to the following dual program:

$$\min_{l_{kj}, k=1\ldots|S|, j=1\ldots|S|} \sum_{k,j=1}^{|S|} l_{kj} d(s_k, s_j)$$

subject to: $\forall k. \; \sum_j l_{kj} = P(s_k)$
$\forall j. \; \sum_k l_{kj} = Q(s_j)$
$\forall k, j. \; l_{kj} \geq 0$

The origins of the Kantorovich metric lie in mass transportation theory. Consider two copies of the state space, one in which states are labeled as supply nodes, and the other in which states are labeled as demand nodes. Each supply node has a supply whose value is equal to the probability mass of the corresponding state under $P$. Each demand has a value equal to the probability mass of the corresponding state under $Q$. Furthermore, imagine there is a transportation arc from each supply node to each demand node, labeled with a cost equal to the distance of the corresponding states under $d$. This constitutes a transportation network. A flow with respect to this network is an assignment of quantities to be shipped along each arc subject to the conditions that the total flow leaving a supply node is equal to its supply, and the total flow entering a demand node is equal to its demand. The cost of a flow along an arc is the value of the flow along that arc multiplied by the cost assigned to that arc. The goal of the Kantorovich optimal mass transportation problem is to find the best total flow for the given network, i.e. the flow of minimal cost. This formulation is captured exactly in the dual program above. The distance assigned to $P$ and $Q$, $T_K(d)(P,Q)$, is the cost of the optimal flow, which is known to be computable in strongly polynomial time. This formulation can be computed in $O(|S|^2 \log |S|)$ time (Orlin, 1988).

---

[2] Note that the Kullbach-Leibler divergence, also known as KL-distance, which is commonly used to estimate the similarity of probability distributions, is *not* a metric.

Since the underlying cost function $d$ is a semimetric, the Kantorovich metric may be further simplified.

**Lemma 3.1.** *Let $d \in \mathcal{M}$. Then*

$$T_K(d)(P,Q) = \max_{v_C} \sum_{C \in S/R_d} (P(C) - Q(C)) v_C$$

$$\text{subject to: } \forall C, D. \ v_C - v_D \leq \min_{i \in C, j \in D} d(s_i, s_j)$$

$$\forall C. \ 0 \leq v_C \leq 1$$

*and $T_K(d)(P,Q) = 0 \Leftrightarrow P(C) = Q(C), \forall C \in S/R_d$.*

*Proof.* Let $\{v_i\}$ be any feasible solution to the primal LP for $T_K(d)(P,Q)$. Note that if $s_i R_d s_j$ then we must have $v_i = v_j$. Define for each $C \in S/R_d$, $v_C = v_i$ for some $s_i \in C$. Then collecting terms yields the desired expression. From this expression it is clear that if $P(C) = Q(C)$ for every equivalence class $C$, then $T_K(d)(P,Q) = 0$. For the converse, suppose that $\exists C$ such that $P(C) \neq Q(C)$. Without loss of generality, suppose $P(C) > Q(C)$. Clearly $C \neq S$, so we may take $v_C = \min_{k \in C, j \in S \setminus C} d(s_k, s_j)$ and $v_D = 0$ for all other classes and obtain a positive lower bound on $T_K(d)(P,Q)$. □

By contrast, the total variation probability metric, $T_{TV}$, is defined independently of $d$ by

$$T_{TV}(P,Q) = \frac{1}{2} \sum_{s \in S} |P(s) - Q(s)|$$

which is half the $L^1$-norm of $P$ and $Q$. It clearly has the advantage of being simply defined and easily computable. Yet, it may still be placed within the previous context since $T_{TV}$ can be expressed as $T_K(\mathbb{I}_{[X \neq Y]})$.

## 4 Bisimulation Metrics

Our construction of bisimulation metrics is heavily based on the following two lemmas, which are important consequences of lemma 3.1. Here the usefulness of the Kantorovich metric becomes evident.

**Lemma 4.1.** *If $d$ is a bisimulation metric then $\forall s, s' \in S$,*

$$d(s,s') = 0 \Leftrightarrow \forall a \in A. \ (r_s^a = r_{s'}^a \text{ and } T_K(d)(P_s^a, P_{s'}^a) = 0) \tag{1}$$

Since condition 1 is necessary for $d \in \mathcal{M}$ to be a bisimulation metric, the question naturally arises as to whether or not it is sufficient as well. In general, the answer is negative. However, it is sufficient for $d$ to be a bisimulation relation metric.

**Lemma 4.2.** *Suppose $d \in \mathcal{M}$ satisfies (1). Then*

$$d(s,s') = 0 \Rightarrow s \sim s'$$

We have stated that our goal is to construct bisimulation metrics that provide useful information concerning the optimal values of states, but we have not mentioned how this can be done. For inspiration we look to the Bellman optimality equations for the optimal value function, which yield the following bound:

$$|V^*(s) - V^*(s')| \leq \max_{a \in A} (|r_s^a - r_{s'}^a| + \gamma | \sum_{u \in S} (P_{su}^a - P_{s'u}^a) V^*(u)|)$$

The first component of the RHS is simply the distance in immediate rewards, while the second component is strikingly similar to the primal LP for the Kantorovich distance in distributions.

Based on these observations we fix a particular form for our bisimulation metrics, namely

$$d(s,s') = \max_{a \in A} (c_R |r_s^a - r_{s'}^a| + c_T d_P(P_s^a, P_{s'}^a))$$

where $d_P$ is some probability metric and $c_R$ and $c_T$ are two positive 1-bounded constants. Intuitively, these constants weight the importance given to the distance between rewards relative to the distance between transition probabilities respectively. For instance, in MDPs a natural choice would be $c_T = \gamma$ and $c_R = (1 - \gamma)$. The particular choice of probability metric leads to two kinds of bisimulation metrics, which we now describe in detail.

### 4.1 Fixed-Point Metrics

In this section, we will use the Kantorovich distance as a basis for formulating a bisimulation metric. Before we do so, we need some definitions and results from fixed-point theory. These may be found, for example, in (Winskel, 1993). We present them in general notation first, then we explain it in the context of our problem.

Let $(X, \preceq)$ be a partial order. An ω-*chain* of this partial order is an increasing sequence $\{x_n\}$. The partial order is said to be an ω-*complete* partial order (ω-cpo) if it contains least upper bounds of all ω-chains. It is called an ω-*cpo with bottom* if it additionally contains a least element, $\bot$, called *bottom*. A function $f : X \to Y$ between ω-cpos is said to be *monotonic* if $x \preceq x' \Longrightarrow f(x) \preceq f(x')$. It is *continuous* if for every ω-chain $\{x_n\}$, $f(\sqcup_{n \in \mathbb{N}} \{x_n\}) = \sqcup_{n \in \mathbb{N}} \{f(x_n)\}$. A point $x \in X$ is said to be a *prefixed-point* of $f$ if $f(x) \preceq x$. It is a *fixed-point* if $x = f(x)$. With these definitions, the following important theorem can be established.

**Theorem 4.3 (Fixed-Point Theorem).** *Let $f : X \to X$ be a continuous function on an ω-cpo with bottom $X$. Define $fix(f) = \sqcup_{n \in \mathbb{N}} f^n(\bot)$. Then $fix(f)$ is the least prefixed-point of $f$ and the least fixed-point of $f$.*

In order to use this result, we equip $\mathcal{M}$ with the usual pointwise ordering: $d \leq d'$ iff $d(s,s') \leq d'(s,s')$ for all $s,s' \in S$. As a result, we obtain an $\omega$-cpo with bottom, where $\bot$ is the constant zero function and $\sqcup_{n \in \mathbb{N}}\{d_n\}$ is given by $\sqcup_{n \in \mathbb{N}}\{d_n\}(s,s') = \sup_{n \in \mathbb{N}}\{d_n(s,s')\}$. Moreover, the same can be said of the set $\mathcal{M}_P$ of semimetrics on the set of probability functions on $S$. With this in mind it is now easy to see that this ordering is preserved by the Kantorovich metric, i.e.

**Lemma 4.4.** $T_K : \mathcal{M} \to \mathcal{M}_P$ is continuous.

*Proof.* See appendix. □

We are now ready to establish the bisimulation metric based on the Kantorovich probability metric:

**Theorem 4.5.** Let $c_R$, $c_T \geq 0$ with $c_R + c_T \leq 1$. Define $F: \mathcal{M} \to \mathcal{M}$ by

$$F(d)(s,s') = \max_{a \in A}(c_R|r_s^a - r_{s'}^a| + c_T T_K(d)(P_s^a, P_{s'}^a))$$

Then $F$ has a least fixed-point, $d_{fix}$, and $d_{fix}$ is a bisimulation metric.

*Proof.* Clearly, existence of the least fixed-point will follow from theorem 4.3, so we only need to show that $F$ is continuous. For future reference we will denote the iterates, $F^n(\bot)$, by $d_n$ and remark that they form an $\omega$-chain in $\mathcal{M}$.

Continuity of $F$ follows from lemma 4.4, since it establishes the monotonicity of $F$, and from the fact that given an $\omega$-chain $\{x_n\}$ in $\mathcal{M}$ and a pair of states $s$ and $s'$,

$$F(\sqcup_{n \in \mathbb{N}}\{x_n\})(s,s')$$
$$= \max_{a \in A}(c_R|r_s^a - r_{s'}^a| + c_T T_K(\sqcup_{n \in \mathbb{N}}\{x_n\})(P_s^a, P_{s'}^a))$$
$$= \max_{a \in A}(c_R|r_s^a - r_{s'}^a| + c_T \sup_{n \in \mathbb{N}}\{T_K(x_n)(P_s^a, P_{s'}^a)\})$$
$$= \sup_{n \in \mathbb{N}} \max_{a \in A}(c_R|r_s^a - r_{s'}^a| + c_T T_K(x_n)(P_s^a, P_{s'}^a))$$
$$= \sup_{n \in \mathbb{N}}\{F(x_n)(s,s')\} = (\sqcup_{n \in \mathbb{N}}\{F(x_n)\})(s,s')$$

So $d_{fix}$ exists, and $d_{fix} = \sqcup_{n \in \mathbb{N}} F^n(\bot)$. Note that by construction, $d_{fix}$ satisfies (1), and so, from lemma 4.2 $d_{fix}(s,s') = 0 \Rightarrow s \sim s'$. On the other hand, since $\mathbb{I}_{[X \not\sim Y]}$ is a bisimulation metric, by applying lemma 4.1 and the definition of $F$, $F(\mathbb{I}_{[X \not\sim Y]})$ is also a bisimulation metric. Therefore, $F(\mathbb{I}_{[X \not\sim Y]}) \leq \mathbb{I}_{[X \not\sim Y]}$, i.e. $\mathbb{I}_{[X \not\sim Y]}$ is a prefixed-point of $F$. So $d_{fix} \leq \mathbb{I}_{[X \not\sim Y]}$, since $d_{fix}$ is the least prefixed-point of $F$. Thus, $s \sim s' \implies d_{fix}(s,s') = 0$. □

Note that by induction $d_{fix} - d_n \leq c_T^n$ for every $n$. Thus, we can compute $d_{fix}$ up to a prescribed degree of accuracy $\delta$ by iteratively applying $F$ for $\lceil \frac{\ln \delta}{\ln c_T} \rceil$ steps. Since this essentially reduces to computing a Kantorovich metric at each iteration for every action and pair of states, $d_{fix}$ can be computed in $O(|A||S|^4 \log |S| \frac{\ln \delta}{\ln c_T})$ operations.

### 4.2 Metrics based on Total Variation

We remarked in the proof of theorem 4.5 that $F(\mathbb{I}_{[X \not\sim Y]})$, which we now denote by $d_\sim$ is also a bisimulation metric. The advantage to using $d_\sim$ in place of $d_{fix}$ is that its component probability semimetric, $T_K(\mathbb{I}_{[X \not\sim Y]})$, admits an explicit, easily computable formulation, similar to that of the total variation metric.

**Lemma 4.6.**

$$T_K(\mathbb{I}_{[X \not\sim Y]})(P,Q) = \frac{1}{2}\sum_{C \in S/\sim}|P(C) - Q(C)|$$

*Proof:* By lemma 3.1, we have:

$$T_K(\mathbb{I}_{[X \not\sim Y]})(P,Q) = \max_{u_C} \sum_{C \in S/\sim}(P(C) - Q(C))u_C$$

subject to: $\forall C, D. \; u_C - u_D \leq \min_{i \in C, j \in D} \mathbb{I}_{[X \not\sim Y]}(s_i, s_j)$

$$\forall C. \; 0 \leq u_C \leq 1$$

However, for distinct bisimulation equivalence classes $C$ and $D$, $\mathbb{I}_{[X \not\sim Y]}(s_i, s_j)$ is 1, and so the first constraint is extraneous. Thus, if we define $u_C$ to be 1 if $P(C) \geq Q(C)$ and 0 otherwise, then it is clear that $\{u_C\}$ is a feasible solution at which the maximum is achieved. For this solution we have,

$$T_K(\mathbb{I}_{[X \not\sim Y]})(P,Q) = \sum_{C \in S/\sim}(P(C) - Q(C))u_C$$
$$= \sum_{C \in S/\sim}(P(C) - Q(C))(u_C - \frac{1}{2}) + \frac{1}{2}(P(S) - Q(S))$$
$$= \frac{1}{2}\sum_{C \in S/\sim}|P(C) - Q(C)| \qquad \square$$

Thus, $d_\sim$ can be computed via the bisimulation partition in $O(|A||S|^3)$ operations.

## 5 Value Function Bounds

We are now ready to provide value function bounds. We will state the bounds in terms of $d_{fix}$ only. The bounds hold immediately for $d_\sim$ as well, because $d_{fix} \leq d_\sim$.

**Theorem 5.1.** Suppose $\gamma \leq c_T$. Then $\forall s, s' \in S$:

$$c_R|V_n(s) - V_n(s')| \leq d_n(s,s')$$
$$c_R|V^*(s) - V^*(s')| \leq d_{fix}(s,s')$$

*Proof:* Clearly the proof of the second item follows from the first by taking limits. For the proof of the first item we proceed by induction. Note that since $\gamma \leq c_T$

$$0 \leq \frac{c_R \gamma}{c_T} V_i(u) \leq \frac{(1-c_T)\gamma}{c_T(1-\gamma)} \leq 1$$

and by the induction hypothesis

$$\frac{c_R\gamma}{c_T}V_n(u) - \frac{c_R\gamma}{c_T}V_n(v) \leq c_R|V_n(u) - V_n(v)| \leq d_n(u,v).$$

So $\{\frac{c_R\gamma}{c_T}V_n(u) : u \in S\}$ constitutes a feasible solution to the primal LP for $T_K(d_n)(P_s^a, P_{s'}^a)$. It follows that

$$\begin{aligned}
&c_R|V_{n+1}(s) - V_{n+1}(s')| \\
&= c_R|\max_{a \in A}(r_s^a + \gamma \sum_{u \in S} P_{su}^a V_n(u)) - \max_{a \in A}(r_{s'}^a + \gamma \sum_{u \in S} P_{s'u}^a V_n(u))| \\
&\leq c_R \max_{a \in A}|r_s^a - r_{s'}^a + \gamma \sum_{u \in S}(P_{su}^a - P_{s'u}^a)V_n(u)| \\
&\leq \max_{a \in A}(c_R|r_s^a - r_{s'}^a| + c_T|\sum_{u \in S}(P_{su}^a - P_{s'u}^a)\frac{c_R\gamma}{c_T}V_n(u)|) \\
&\leq \max_{a \in A}(c_R|r_s^a - r_{s'}^a| + c_T T_K(d_n)(P_s^a, P_{s'}^a)) \\
&= F(d_n)(s,s') = d_{n+1}(s,s') \qquad \square
\end{aligned}$$

These bounds can be extended to relate the optimal values of states in the given MDP and an aggregate MDP. First, let us fix some notation and assumptions concerning the form of an aggregate MDP. We assume the aggregate is given by $(S', A, \{P_{CD}^a : a \in A, C, D \in S'\}, \{r_C^a : a \in A, C \in S'\})$ where $S'$ is a partition of the state space $S$, $A$ is the same finite set of actions, and transition probabilities and rewards are each averaged over equivalence classes, i.e.

$$P_{CD}^a = \frac{1}{|C|}\sum_{s \in C} P_s^a(D) \text{ and } r_C^a = \frac{1}{|C|}\sum_{s \in C} r_s^a$$

Additionally in the following we will denote the map from $S$ to $S'$ taking a state to its equivalence class by $\rho$, and the average distance from state $s$ to all states in its equivalence class under semimetric $d$, by $g(s,d) = \frac{1}{|\rho(s)|}\sum_{s' \in \rho(s)} d(s,s')$.

**Theorem 5.2.** *Suppose $\gamma \leq c_T$. Then $\forall s \in S$, the following inequalities hold:*

$$c_R|V_n(\rho(s)) - V_n(s)| \leq g(s,d_n) + \sum_{k=1}^{n-1}\gamma^{n-k}\max_{u \in S}g(u,d_k)$$

$$c_R|V^*(\rho(s)) - V^*(s)| \leq g(s,d_{fix}) + \frac{\gamma}{1-\gamma}\max_{u \in S}g(u,d_{fix}))$$

*Proof:* See appendix.

The proposed distance metrics can be used for aggregating states in a straightforward way. For some positive $\varepsilon$ we choose several seed states and for each, we cluster all the states within an $\varepsilon$-neighborhood (while ensuring that each state is placed in only one cluster). Then for a cluster $C$ and any state $s$ belonging to it, the above theorem tells us that $|V^*(C) - V^*(s)| \leq \frac{2\varepsilon}{c_R(1-\gamma)}$, provided $\gamma \leq c_T$. Thus, as $\varepsilon$ decreases, the optimal values of a class and its states converge.

## 6 Illustration

We illustrate our distance metrics and error bounds on a very simple toy MDP, consisting of a $5 \times 5$ grid. There are 5 actions, north, south, east, west and stay. Transitions for each cell are uniformly distributed among adjacent cells. Rewards are distributed as follows. Moving south from rows 1-4 to rows 2-5 yields rewards of 0.1, 0.2, 0.3, 0.4 respectively. Moving east from columns 1-4 to columns 2-5 yields rewards of 0.5, 0.53, 0.56, 0.59 respectively. Finally, staying in the southeast corner yields a reward of 1. All other actions give 0 reward. We used these parameters in order to be able to inspect the partitions obtained. More extensive (but similar) illustrations, using random MDPs, are discussed in (Ferns, 2003).

In all experiments, $c_R = 1 - \gamma$ and $c_T = \gamma$. We first compute the pairwise distances between all pairs of states. Note that this is not a practical approach; here, we are just trying to understand the behavior of the metrics. Then, from an initial seed state we grow a partition of $\varepsilon$-clusters of states, adding a new cluster each time we encounter a state at distance greater than $\varepsilon$ from the seed states of each cluster presently in the partition. Of course, the quality of the partition will depend on the choice of seeds, and more sophisticated methods can be employed here (e.g., picking the seeds for subsequent partitions as far as possible from the previous ones). Once a partition is established, we perform value iteration to find the value of the optimal policy.

We varied the parameter $\varepsilon$ which bounds the allowed distance between states, and well as the discount factor $\gamma$. Note that low $\varepsilon$ (close to 0) means that we only allow states to be aggregated if they are very close in terms of the distance. Hence, at this end of the spectrum, very little aggregation will occur and the value function in the aggregated MDP should be very close (or identical) to the one in the original MDP. When $\varepsilon = 1$, all states can be aggregated, resulting in a single-state MDP, and a poor approximation of the optimal value function. Figure 2 shows the size of the aggregated MDPs, obtained using the Kantorovich metric and the total variation metric, for values of $\gamma = 0.1$, 0.5 and 0.9. The two metrics are close for low $\gamma$ but behave quite differently for high values of $\gamma$ (which are typical in the MDP community). In particular, the total variation metric has a very abrupt transition from no aggregation to aggregating all states in one lump. We note, though, that this metric is much faster to compute (by an order of $100 - 10000$ in our experiments, in a Java implementation). Figure 3 compares the metrics in terms of actual and estimated error. The lower curves represent the maximum error between the optimal value functions of the aggregated MDP and the original MDP. The higher curves are the upper bound on the error, based on Theorem 5.2. The straight line is the naive estimate, $\frac{2\varepsilon}{c_R(1-\gamma)}$. Note that the bounds in the theorem are much tighter than the naive bound (which is omitted in the last graph to make the fig-ure clear). The bounds get looser as $\gamma$ increases, due to the $\frac{1}{1-\gamma}$ factor. We note, though, that the shape of the bound mimics very well the shape of the actual error.

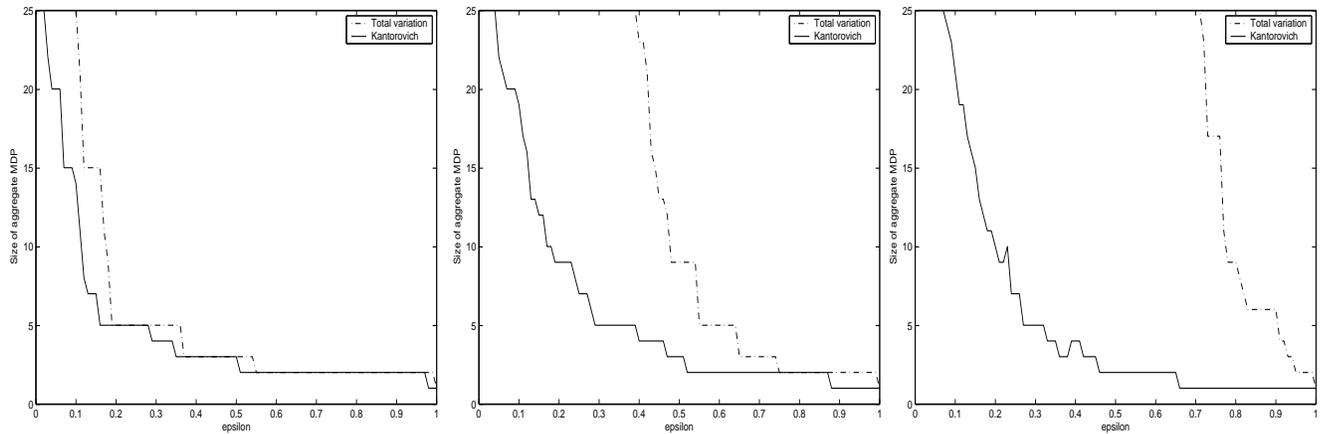

Figure 2: Size of aggregated MDP as a function of $\varepsilon$, for $\gamma = 0.1$ (left), $\gamma = 0.5$ (middle) and $\gamma = 0.9$ (right).

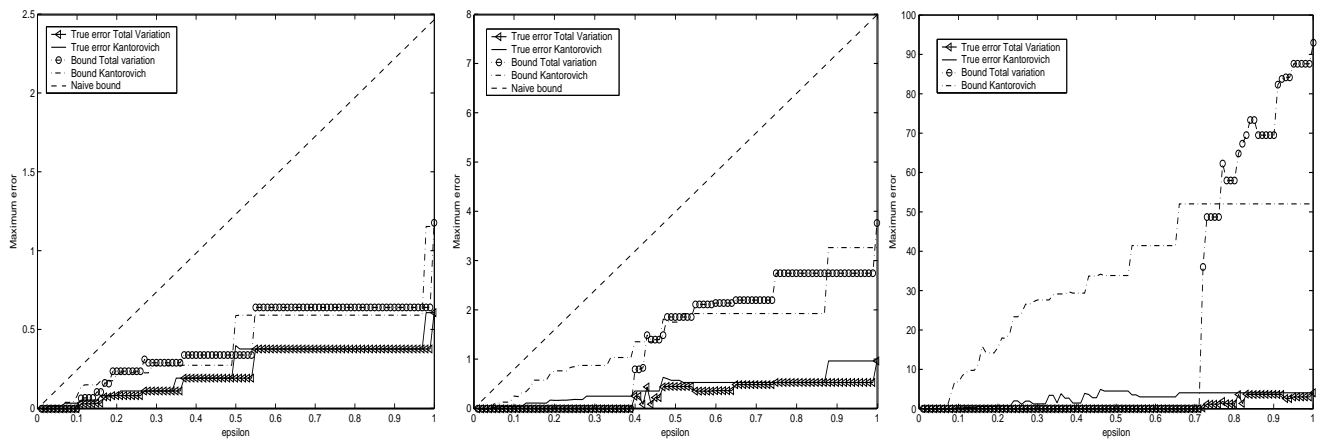

Figure 3: True error and estimated error bounds between the optimal value function of the original and aggregated MDP, as a function of $\varepsilon$, for $\gamma = 0.1$ (left), $\gamma = 0.5$ (middle) and $\gamma = 0.9$ (right).

## 7 Conclusion

In this paper, we introduced metrics for measuring the distance between the states of an MDP, based on the notion of bisimulation. Unlike equivalence relations, the metrics are robust to perturbations in the parameters of the MDP: if two bisimilar states are slightly perturbed, the metric will still show them as "close". Moreover, the same can be said of the states' optimal values, as reflected by the bounds relating these to our metrics. Such metrics are obviously useful for state aggregation, but also for other value function approximators (e.g. memory-based). We are currently pursuing an interesting connection to diffusion kernels on graphs (Kondor & Lafferty, 2002).

The existence of bisimulation metrics for finite MDPs allows us to tackle compression of such systems in a new manner. A metric defined on the state space of an MDP can be extended to a metric on the space of finite MDPs. With this in mind, we are now concerned with answering the following question: given a finite MDP and a positive integer $k$, what is its "best" $k$-state approximation? Here by "best" we mean a $k$-state MDP of minimal distance to the original. We also aim to extend these results to other probabilistic models. We have mostly established an extension for continuous-state MDPs. In the future, we hope to tackle factored MDPs and partially observable MDPs as well.

## Appendix: Proof of Lemma 4.4

Fix probability functions $P$ and $Q$. Monotonicity of $T_K$ follows from the primal LP: for, if $d \leq d'$ then every feasible solution to $T_K(d)(P,Q)$ is a feasible solution to $T_K(d')(P,Q)$. Thus, $T_K(d) \leq T_K(d')$.

Next, given $\omega$-chain $\{d_n\}$ note that by monotonicity $\sup T_K(d_n)(P,Q) \leq T_K(\sqcup d_n)(P,Q)$. For the other direction, we use the dual LP. For each $n$, let $\{l_{kj}^{(n)}\}$ denote a feasible solution of $T_K(d_n)$ yielding the minimum. Then each is also a feasible solution for $T_K(\sqcup d_n)$. Define $\varepsilon_{kj}^{(n)} = (\sqcup\{d_n\})(s_k, s_j) - d_n(s_k, s_j)$ and $\delta_{kj} = \min(P(s_k), Q(s_j))$. Then for every $k$, $j$, and $n$, $\varepsilon_{kj}^{(n)} \geq 0$, $\lim_{n \to \infty} \varepsilon_{kj}^{(n)} = 0$, and $l_{kj}^{(n)} \leq \delta_{kj}$. Thus,

$$T_K(\sqcup\{d_n\})(P,Q) \leq \sum_{k,j} l_{kj}^{(n)}(\sqcup\{d_n\})(s_k, s_j)$$

$$\leq T_K(d_n)(P,Q) + \sum_{k,j} l_{kj}^{(n)} \varepsilon_{kj}^{(n)} \leq \sup T_K(d_n)(P,Q) + \sum_{k,j} \delta_{kj} \varepsilon_{kj}^{(n)}$$

By taking $n \to \infty$ on both sides of the inequality, we obtain the desired result.

**Appendix: Proof of Theorem 5.2**

Once more we proceed by induction.

$|V_{n+1}(\rho(s)) - V_{n+1}(s)|$
$= |\max_{a \in A}(r_{\rho(s)}^a + \gamma \sum_{D \in S'} P_{\rho(s)D}^a V_n(D)) - \max_{a \in A}(r_s^a + \gamma \sum_{u \in S} P_{su}^a V_n(u))|$

$\leq \frac{1}{|\rho(s)|} \sum_{s' \in \rho(s)} \max_{a \in A}(|r_{s'}^a - r_s^a| +$
$\quad + \gamma |\sum_{D \in S'} \sum_{u \in D} P_{s'u}^a V_n(D) - \sum_{u \in S} P_{su}^a V_n(u)|)$

$\leq \frac{1}{|\rho(s)|} \sum_{s' \in \rho(s)} \max_{a \in A}(|r_{s'}^a - r_s^a| + \gamma |\sum_{u \in S} P_{s'u}^a V_n(\rho(u)) - P_{su}^a V_n(u)|)$

$\leq \frac{1}{|\rho(s)|} \sum_{s' \in \rho(s)} \max_{a \in A}(|r_{s'}^a - r_s^a| + \gamma |\sum_{u \in S} (P_{s'u}^a - P_{su}^a) V_n(u)|$
$\quad + \gamma |\sum_{u \in S} P_{s'u}^a (V_n(\rho(u)) - V_n(u))|)$

$\leq \frac{c_R^{-1}}{|\rho(s)|} \sum_{s' \in \rho(s)} \max_{a \in A}(c_R|r_s^a - r_{s'}^a| + c_T |\sum_{u \in S} (P_{s'u}^a - P_{su}^a) \frac{c_R \gamma}{c_T} V_n(u)|)$
$\quad + \frac{\gamma}{|\rho(s)|} \sum_{s' \in \rho(s)} \max_{a \in A} \sum_{u \in S} P_{s'u}^a |V_n(\rho(u)) - V_n(u)|$

Note by theorem 5.1 that $\{\frac{c_R \gamma}{c_T} V_n(u) : u \in S\}$ constitutes a feasible solution to the primal LP for $T_K(d_n)(P_s^a, P_{s'}^a)$. Hence we can continue as follows:

$\leq \frac{c_R^{-1}}{|\rho(s)|} \sum_{s' \in \rho(s)} \max_{a \in A}(c_R|r_s^a - r_{s'}^a| + c_T T_K(d_n)(P_s^a, P_{s'}^a))$
$\quad + \frac{\gamma}{|\rho(s)|} \sum_{s' \in \rho(s)} \max_{a \in A} \sum_{u \in S} P_{s'u}^a \max_{u \in S} |V_n(\rho(u)) - V_n(u)|$

$\leq \frac{c_R^{-1}}{|\rho(s)|} \sum_{s' \in \rho(s)} d_{n+1}(s, s') + \gamma \max_{u \in S} |V_n(\rho(u)) - V_n(u)|$

$\leq \frac{g(s, d_{n+1})}{c_R} + \gamma \max_{u \in S}(\frac{g(u, d_n)}{c_R} + \sum_{k=1}^{n-1} \gamma^{n-k} \max_{v \in S} g(v, d_k)))$

$\leq \frac{1}{c_R}(g(s, d_{n+1}) + \gamma \max_{u \in S} g(u, d_n) + \sum_{k=1}^{n-1} \gamma^{n+1-k} \max_{u \in S} g(u, d_k))$

$\leq \frac{1}{c_R}(g(s, d_{n+1}) + \sum_{k=1}^{n} \gamma^{(n+1)-k} \max_{u \in S} g(u, d_k))$ □